\documentclass[a4paper,fleqn]{cas-dc}

\usepackage{makecell}
\usepackage{amssymb}
\usepackage{bbding}
\usepackage[numbers]{natbib}
\usepackage{cleveref}
\crefname{figure}{Figure}{Figures}
\crefname{table}{Table}{Tables}
\usepackage{subfig}
\def\tsc#1{\csdef{#1}{\textsc{\lowercase{#1}}\xspace}}
\tsc{WGM}
\tsc{QE}

\begin{document}
\let\WriteBookmarks\relax
\def\floatpagepagefraction{1}
\def\textpagefraction{.001}

\shorttitle{ECT:Fine-grained \underline{E}dge Detection with Learned \underline{C}ause \underline{T}okens}    

\shortauthors{Shaocong}

\title [mode = title]{ECT: Fine-grained \underline{E}dge Detection with Learned \underline{C}ause \underline{T}okens  }



%

\author[1,2]{Shaocong Xu}
\ead{xushaocong@stu.xmu.edu.cn}
\author[2,3]{Xiaoxue Chen}
\ead{chenxiaoxue@air.tsinghua.edu.cn}
\author[4]{Yuhang Zheng}
\ead{zyh_021@buaa.edu.cn}
\author[2]{Guyue Zhou}
\ead{zhouguyue@air.tsinghua.edu.cn}
\author[6]{Yurong Chen}
\ead{yurong.chen@intel.com}
\author[5]{Hongbin Zha}
\ead{zha@cis.pku.edu.cn}
\author[2]{Hao Zhao}
\ead{zhaohao@air.tsinghua.edu.cn}
\cormark[1]







\affiliation[1]{organization={School of Informatics, Xiamen University},
            city={Xiamen},
            postcode={361005}, 
            country={China}}

\affiliation[2]{organization={Institute for AI Industry Research (AIR), Tsinghua University},city={Beijing},postcode={100084},country={China}}

\affiliation[3]{organization={Department of Computer Science and Technology, Tsinghua University},city={Beijing},postcode={100084},country={China}}

\affiliation[4]{organization={School of Mechanical Engineering and Automation, Beihang University},city={Beijing},postcode={100084},country={China}}

\affiliation[5]{organization={School of Electronic Engineering and Computer Science, Peking University},city={Beijing},postcode={100084},country={China}}

\affiliation[6]{organization={Intel Labs},city={Beijing},postcode={100026},country={China}}





\cortext[1]{Corresponding author}



\begin{abstract}
In this study, we tackle the challenging fine-grained edge detection task, which refers to predicting specific edges caused by reflectance, illumination, normal, and depth changes, respectively. Prior methods exploit multi-scale convolutional networks, which are limited in three aspects: (1) Convolutions are \textbf{local} operators while identifying the cause of edge formation requires looking at far away pixels. (2) Priors specific to edge cause are \textbf{fixed} in prediction heads. (3) Using separate networks for generic and fine-grained edge detection, and the constraint between them may be \textbf{violated}. To address these three issues, we propose a two-stage transformer-based network sequentially predicting generic edges and fine-grained edges, which has a global receptive field thanks to the attention mechanism. The prior knowledge of edge causes is formulated as four learnable cause tokens in a cause-aware decoder design. Furthermore, to encourage the consistency between generic edges and fine-grained edges, an edge aggregation and alignment loss is exploited. We evaluate our method on the public benchmark BSDS-RIND and several newly derived benchmarks, and achieve new state-of-the-art results. Our code, data, and models are publicly available at \url{https://github.com/Daniellli/ECT.git}.
\end{abstract}




\maketitle

\section{Introduction}
\label{sec:Introduction}

Generic Edges (GEs) detection \cite{canny}\cite{perona1990scale} is one of the most important computer vision topics, which benefits a lot of applications nowadays, like ego pose estimation \cite{robotics1}, target pose estimation \cite{robotics2} and map construction \cite{robotics3}. However, the causes of edge formation can be varied, and confusion between them is potentially harmful for downstream tasks. As such \textbf{fine-grained edge detection} \cite{rindnet} further categorizes edges into four types according to the cause, namely reflectance, illumination, normal, and depth discontinuity respectively. 

Depth Edges (DEs) are sometimes defined as occlusion boundaries \cite{hoiem2011occlusion}\cite{wang2020occlusion} and used to improve depth estimation \cite{using_depth_map_locate_occlusion_edge_and_recons}; Illumination Edges (IEs) detection is a prerequisite for some shadow removal methods \cite{shadow_edge_application}; Normal Edges (NEs) reveal important cues about the orientation of scene elements \cite{recovering_spatial_layout}\cite{schwing2013}; Reflectance Edges (REs) can be used to facilitate drone decision making \cite{wang2021reflectance}.

\begin{figure}[t]
\centerline{\includegraphics[width=1\columnwidth]{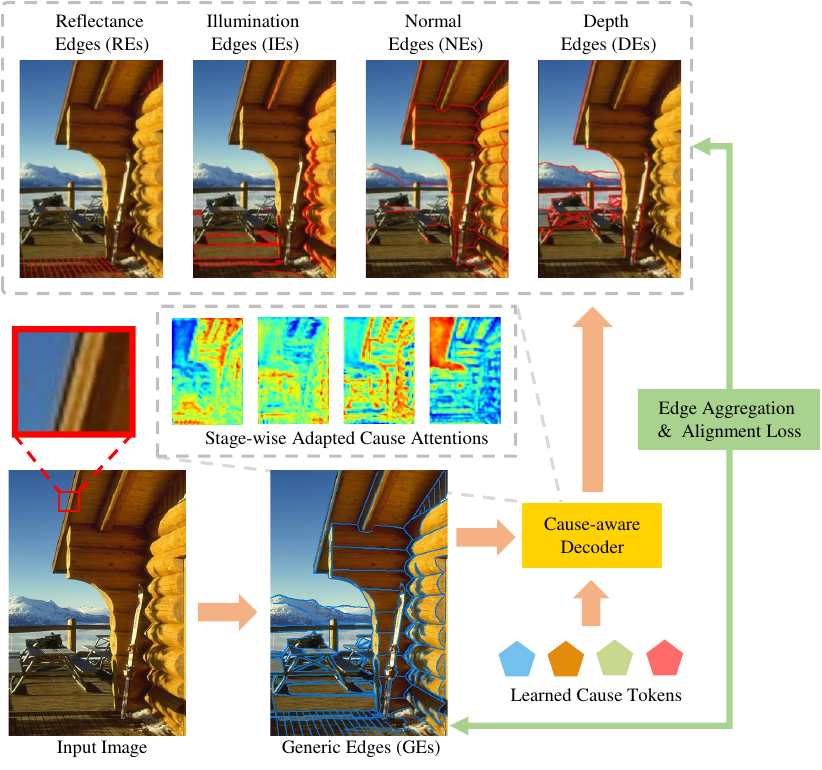}}
\caption{Our method has \textbf{two stages}. The input image is passed through the first stage to generate generic edges. Afterwards, features used for the prediction of generic edges are passed through the second stage (cause-aware decoder) to predict fine-grained edges. \textbf{Learned cause tokens} serve as data-driven priors. \textbf{Edge aggregation and alignment} enforce the consistency between generic edges and fine-grained edges.}
\label{fig:teaser}
\end{figure}

\begin{figure*}[t]
\centerline{\includegraphics[width=1\textwidth]{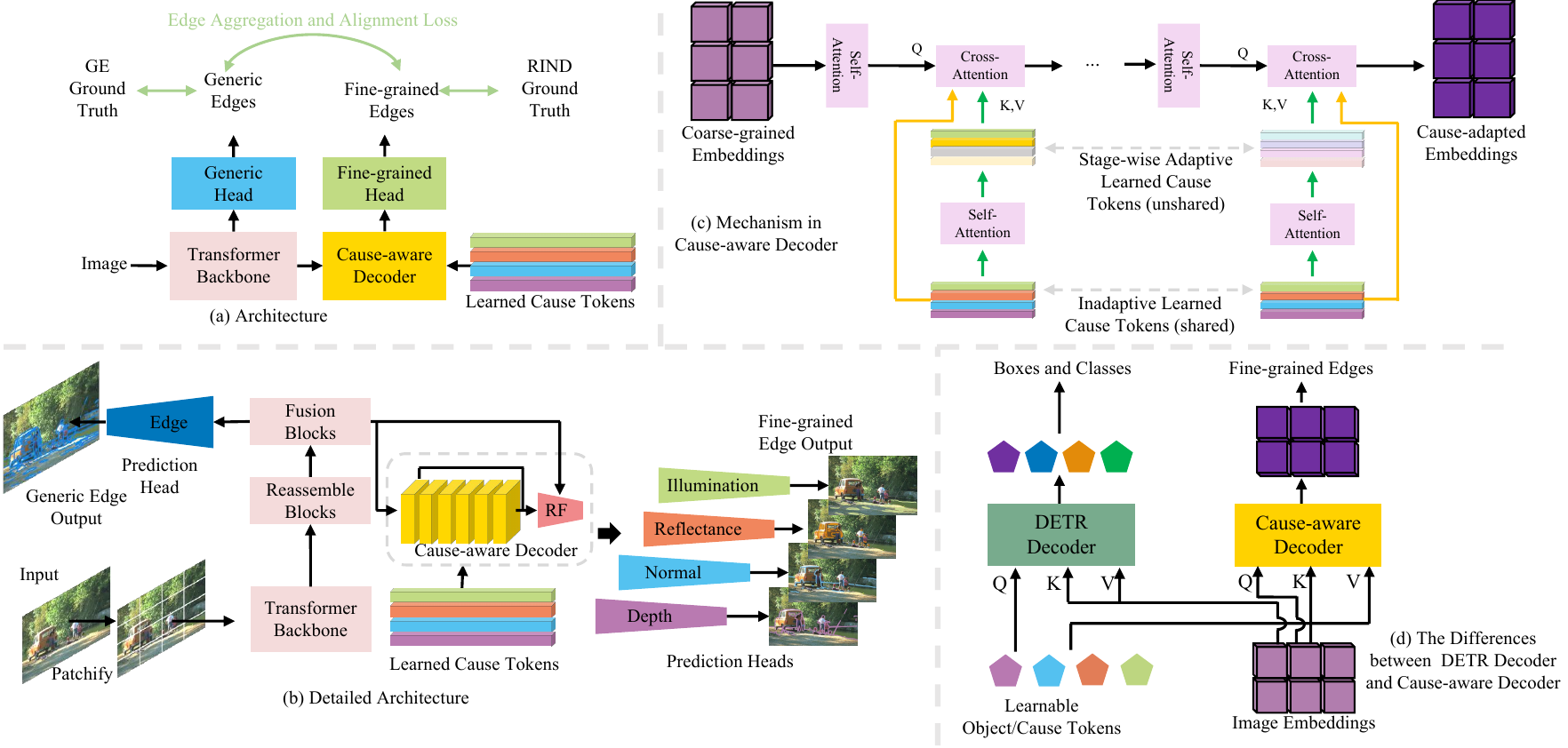}}
\caption{ (a), (b)  and (c) are high-level, detailed network architecture and mechanism in the cause-aware decoder while (d) is one of the main differences between the DETR decoder and the cause-aware decoder; Initially, the image is processed through the transformer backbone, reassembled blocks, and fusion blocks, in addition to the generic edge head for detecting generic edges. Subsequently, the output of the fusion blocks undergoes cause-aware decoding and is further processed by the fine-grained head to detect fine-grained edges.}
\label{fig:main}
\end{figure*}

While the state-of-the-art method RINDNet \cite{rindnet} can produce promising results, it has several intrinsic limitations. \textbf{Firstly}, its multi-scale representations are extracted by convolutions, which are local operators while identifying the cause of edge formation requires looking at far away pixels. As shown by the red box in \cref{fig:teaser}, even humans can struggle to tell that the edge is caused by sky-house depth discontinuity if he/she looks at only the local patch. \textbf{Secondly}, during model training, the prior knowledge about reflection or depth discontinuity accumulates in the weights of prediction heads, which remain fixed during inference. This \textbf{fixed} paradigm has been proven sub-optimal by several recent dense prediction studies \cite{zhang2021k,cheng2021per,strudel2021segmenter}. \textbf{Thirdly}, RINDNet exploits separate networks for generic edge and fine-grained edge detection without incorporating explicit mutual regularization between them. As such, the fact that any fine-grained edge output should be a subset of the generic edge output may be violated during inference.

In order to resolve the aforementioned three issues, we propose a method named ECT, short for \textbf{E}dge \textbf{C}ause \textbf{T}okens. ECT is a two-stage transformer-based architecture that generates generic edges and fine-grained edges sequentially, as shown in \cref{fig:teaser}. Thanks to the global receptive field of attention blocks, ECT learns feature representations that disambiguate local patches like the one outlined by a red box in \cref{fig:teaser}. Prior knowledge about four different edge causes is explicitly modeled by four learnable edge cause tokens, which adapt to different \textbf{cause-aware decoder} stages and generate \textbf{dynamic} (v.s. fixed) kernels for fine-grained edge prediction. Last but not least, an \textbf{Edge Aggregation and Alignment Loss (EA2 Loss)} enforces the consistency between fine-grained edges and generic edges while respecting potential misalignment between them.

The field of fine-grained edge detection now still suffers from the lack of benchmarking datasets. So we derive several new benchmarks from existing datasets.

Our main contributions can be summarized as follows:
\begin{itemize}
\item We propose a new fine-grained edge detection method ECT that addresses the limitations of RINDNet through (1) global receptive fields to disambiguate local patches, (2) learned cause tokens that dynamically adapt to the input for prediction head kernel generation, (3) an edge aggregation and alignment loss that enforces generic/fine-grained edge consistency.
\item  We derive several new benchmarks for the evaluation of fine-grained edge detection, from existing datasets.
\item Experiments on both BSDS-RIND and our new datasets demonstrate that ECT sets new state-of-the-art (SOTA) results. Codes and models will be released.
\end{itemize}

\section{Related Work}

The field of edge detection has a rich history, with early works dating back more than four decades \cite{sobel}. Here we only highlight some representative works. Early edge detectors, such as Sobel \cite{sobel}, and Canny\cite{canny}, utilize image gradients for edge extraction. Learning-based edge detectors \cite{learning-based-1,learning-based-2,learning-based-3} exploit low-level features such as brightness, color, and texture and train a classifier for edge detection. However, these methods are limited by the representation power of hand-crafted features. In contrast, convolutional neural network (CNN)-based edge detectors \cite{CNN-based-1,CNN-based-2,CNN-based-3} offer powerful automatic feature extraction capabilities, but suffer from the loss of fine details as the CNN layers become deeper. To address this issue, several approaches have been proposed that exploit multi-scale representation to extract finer edge details \cite{multi-scale-based-1,multi-scale-based-2,RCF,multi-scale-based-4,hed,multi-scale-based-6,multi-scale-based-7,multi-scale-based-8,multi-scale-based-9}. However, these methods primarily rely on local intensity variation and do not consider the global context, which can result in noisy edges. Recently, EDTER \cite{transformer-based-edge-detector} has been proposed to integrate a transformer into the edge detector to model global dependencies and improve edge detection performance. However, it lacks specific designs for fine-grained edge detection.


Recently, it has been recognized that GEs detection methods are no longer sufficient for downstream tasks in computer vision. Consequently, researchers have directed their attention towards developing fine-grained edge detection techniques that can better handle complex visual scenes. One such technique has been introduced by \cite{illumination_edge_detection_2}, who integrate GE detection in their approach for extracting shadow edges. Furthermore, \cite{shadow_remove} proposes a mean-teacher architecture for shadow detection that can effectively learn from limited labeled data. Similarly, \cite{pavement_crack_detection} presents an algorithm for detecting pavement cracks (REs), which combines multi-scale feature extraction using feature pyramids with the power of hierarchical boosting networks. \cite{illumination_edge_detection_3} proposes a method for detecting shadow edges based on convolutional neural networks (CNNs).

Moreover, RINDNet \cite{rindnet} is the most recent representative work tackling fine-grained edges, namely REs, IEs, NEs, and DEs. RINDNet has the following features: (1) The fine-grained edges are categorized into two groups based on edge cause, a NEs and DEs group, and an IEs and REs group, which are addressed by specific convolutional neural network branches; (2) Priors specific to the edge cause are fixed in prediction heads. (3) It employs a separate network for generic and fine-grained edge detection. In contrast, our work has the following fundamental differences compared with RINDNet: (1) The architecture is different. We designed a two-stage transformer-based architecture for fine-grained and generic edge detection and used a regularization loss to explicitly model the relationship between fine-grained edges and generic edges instead of two separate networks. (2) The prior knowledge of edge causes is modeled as four learnable cause tokens in a cause-aware decoder design rather than fixed in the prediction head.


\begin{figure}[!t]
\centerline{\includegraphics[width=1\columnwidth]{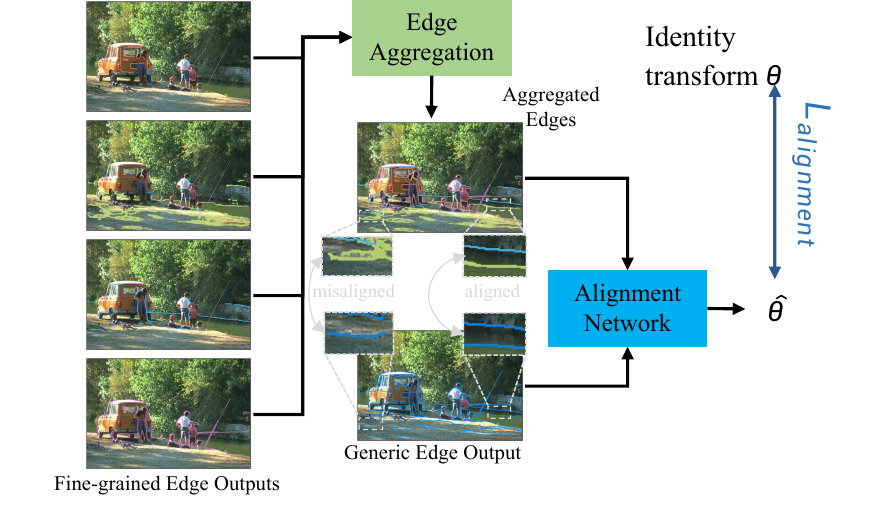}}
\caption{ Pipeline of edge aggregation and alignment loss.}
\label{fig:EA2Loss}
\end{figure}

\section{Method}

The framework of our method is illustrated in \cref{fig:main} (a). The input is an image and the network sequentially predicts generic edges and fine-grained edges. Our insight is that, when addressing the task of fine-grained edge detection, humans tend to start with the easier task (generic edge detection) before thinking about the fine-grained cause. As such, we propose this two-stage architecture. 

Note that directly attaching the fine-grained head to the transformer backbone is also a possible design, but this is hardly different from generic edge detection and makes little use of the domain-specific characteristic of fine-grained edge detection. As such, we propose a novel \textbf{cause-aware decoder with learned cause tokens} to achieve the goal. Using this design, we enforce the learned cause tokens to represent certain intrinsic physical properties of reflectance, illumination, normal, and depth. Furthermore, we enable learned cause tokens to adapt to different cause-aware decoder stages, thus producing dynamic kernels for fine-grained edge prediction.




Moreover, there is a clear relationship between fine-grained edges and generic edges, namely, that the latter is composed of the former. We contend that utilizing this relationship can improve the performance of fine-grained edge detection. Therefore, in addition to supervised loss, we further propose an \textbf{Edge Aggregation and Alignment (EA2) loss} to leverage this relationship and maintain consistency between the fine-grained edge outputs and the generic edge outputs. This approach aggregates fine-grained edges and aligns them with the generic edges, thereby enabling more accurate and reliable edge detection results.

\subsection{Network Architecture}

As illustrated in \cref{fig:main} (b), given an image $\mathcal{X} \in \mathbb{R}^{H \times W \times 3}$,  we first feed it into a transformer-based edge detection backbone, which consists of a ResNet backbone, a ViT encoder, reassemble blocks \cite{cerberus}, fusion blocks \cite{cerberus}, and a prediction head \cite{cerberus}. It generates a generic edge prediction result $\mathcal{E}^e \in \mathbb{R} ^{H \times W \times 1}$ and a high-level feature map $\mathcal{M} \in \mathbb{R}^{ \frac{H}{ s} \times \frac{W}{s} \times D}$ serving as coarse-grained embeddings.

Subsequently, a cause-aware decoder (Sec.~\ref{sec: decoder}) refines these embeddings by incorporating four learnable tokens referred to as learned cause tokens (Sec.~\ref{sec: learned cause token}). Finally, a fine-grained head uses the refined embeddings to make the final prediction of fine-grained edge maps $\mathcal{E}^r, \mathcal{E}^i, \mathcal{E}^n, \mathcal{E}^d \in \mathbb{R}^{H \times W \times 1}$. 


\begin{figure}[!h]
\centerline{\includegraphics[width=1\columnwidth]{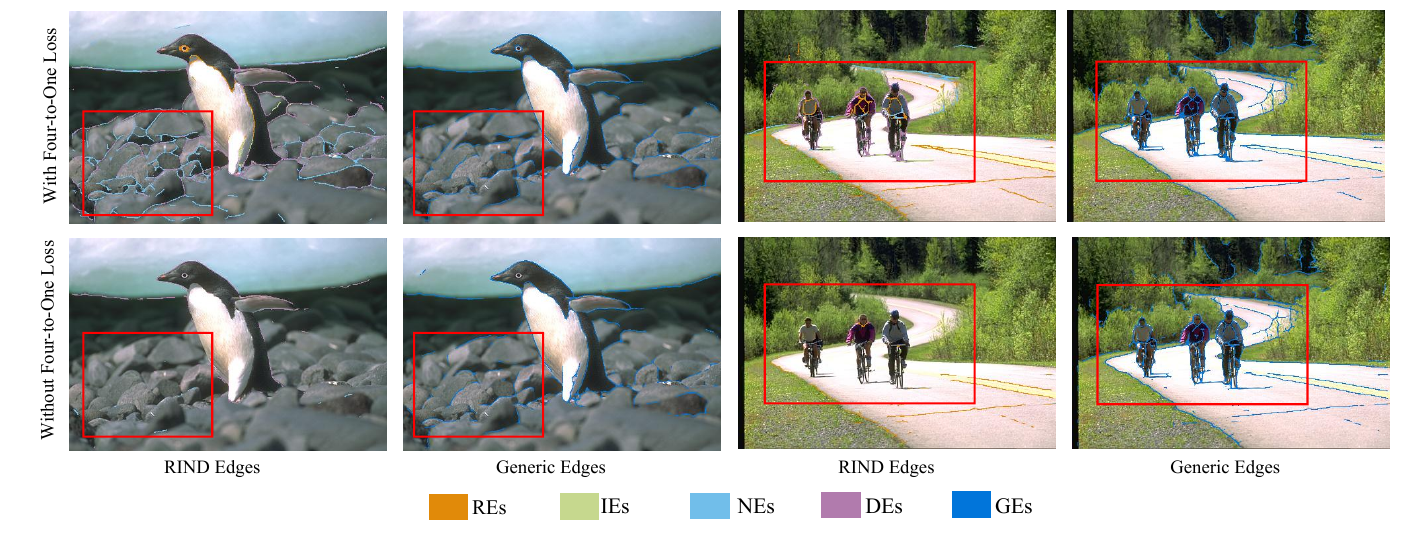}}
\caption{Qualitative results for edge aggregation and alignment loss. The situation of missing edges and the contradiction between fine-grained edges and generic edges can be successfully reduced thanks to the edge aggregation and alignment loss.}
\label{fig:loss_qualitative_results}
\end{figure}

\subsection{Stage-wise Adaptive Learned Cause Tokens}
\label{sec: learned cause token}




Different fine-grained edges with different characteristics are caused by different reasons. Specifically, photometric factors are mostly related to the reflectance edge and the illumination edge. While variations in illumination generate illumination edges (e.g., shadows, light sources, and highlights), changes in the material appearance (such as texture and color) induce reflection edges. Normal edges and depth edges, in contrast, reflect changes in object surfaces' geometry or depth discontinuities. However, in previous method \cite{rindnet}, priors specific to edge causes are learned during the training process but fixed during inference in the prediction head. This formulation fails to adequately account for the differences among fine-grained edges.

Hence, inspired by the learnable object token mechanism of DETR\cite{DETR},  we represent causes of fine-grained edge formation with four learnable tokens, $\mathcal{R}, \mathcal{I}, \mathcal{N}, \mathcal{D} \in \mathbb{R}^{D \times 1}$, which we name as learned cause tokens. 


There are two possible designs, as shown by the flow of the yellow arrow and green arrow in \cref{fig:main} (c). The flow of the yellow arrow illustrates a trivial cause token design in which different transformer decoder stages use the same inadaptive cause tokens as the key and value. This is not an optimal design because different images in different decoder stages need to focus on different regions for different fine-grained cause justifications.

Therefore, we propose our stage-wise adaptive cause token design. The process involves stacking the learned cause tokens and passing them through self-attention \cite{transformer}, generating the stage-wise adaptive learned cause tokens $\psi'$, as illustrated by the flow of green arrows in \cref{fig:main} (c), which can be formulated as follows:

\begin{equation}
\begin{aligned}
\psi &= {\rm Concatenate}(\mathcal{R},\mathcal{I}, \mathcal{N}, \mathcal{D}), \\
\psi' &= {\rm SelfAttention}(\psi,\psi,\psi).
\end{aligned}
\label{eq:stage-wise adapted attention }
\end{equation}

Through this design, specific priors related to edge causes are learned during the training process. During inference, these priors adapt to different cause-aware decoder stages, allowing for the generation of dynamic kernels for fine-grained edge detection.

\subsection{Cause-aware Decoder}
\label{sec: decoder}
The basic idea of learned cause tokens is inspired by the learned object tokens in the DETR method \cite{DETR}, but as shown in \cref{fig:main} (d), they have substantial differences. In both DETR and our decoder, there are image tokens and some learned (object or cause) tokens. In DETR learned object tokens serve as queries while in our decoder image tokens serve as queries. This allows our decoder to give dense predictions in every stage.


Given the coarse-grained tokens $\mathcal{M}$, we first downsample  the $\mathcal{M}$ to $\mathcal{M}' \in \mathbb{R}^{ \frac{H}{ 4 \times s} \times \frac{W}{ 4 \times s} \times D}$. $M'$ are then flattened into the set of $\frac{H}{  4 \times s} \times \frac{W}{  4 \times s}$ tokens denoted as $M''$, each of which is a local image embedding.

Afterward, as shown in \cref{fig:main} (c), we feed $\mathcal{M}''$ into $N$ decoder stages one by one. In the first stage, we feed $\mathcal{M}''$  into self-attention and use it as a query, the stage-wise adaptive learned cause token $\psi'$ as key and value together pass through the cross-attention, generating cause-adapted embedding $\mathcal{M}''_{1}$, which can be formulated as follows: 
\begin{equation}
\begin{aligned}
\mathcal{M}_{\rm sa} &={\rm SelfAttention}(\mathcal{M}{''},\mathcal{M}{''},\mathcal{M}{''}),\\
\mathcal{M}{''_{1}} &= {\rm CrossAttention}(M_{\rm sa},{\psi}',\psi').
\end{aligned}
\label{eq:cause-aware decoder layer}
\end{equation}
The operation in the $n$-th decoder stage involves $\mathcal{M}''_{\text{n-1}}$ and the different stage-wise adaptive learned cause tokens $\psi'$.

As illustrated in \cref{fig:main} (b), after $N$ stage processing, we use a Residual Fusion (RF) module to generate the final cause-adapted image embeddings. 

We first pick up $\mathcal{M}''_{1}$, the output of the $1$-st decoder stage, and $\mathcal{M}''_{N}$, the output of the $N$-th decoder stage, upsampling and merging them into the size of $\mathcal{M}$ and adding them to $\mathcal{M}$ to generate the final cause-adapted image embeddings. The formulation can be expressed as follows:
\begin{equation}
\begin{aligned}
\mathcal{M}_{\text{cause-adapted}} &={\rm UP}({\rm UP}(\mathcal{M}''{_{N}}) + {\rm UP}(\mathcal{M}{''_{1}})) + \mathcal{M}, \\
{\rm UP}(x) &={\rm upsample}({\rm RCU}(x)).
\label{eq:FU}
\end{aligned}
\end{equation}
Here, RCU indicates the residual convolutional unit. Each UP operation represents a $2\times$ upsampling.

\subsection{Edge Aggregation and Alignment Loss}


The relationship between generic edges and fine-grained edges is unequivocal: the latter stem from the former. Nonetheless, conventional approaches to edge detection employ distinct networks for generic and fine-grained edges, overlooking their interdependence. And we believe that leveraging this relationship can improve the performance of fine-grained edge detection by aligning fine-grained edges with generic edges.

However, how to align the fine-grained edges with generic edges is challenging. Fundamental principles do exist: 1). Such an alignment approach should serve four kinds of fine-grained edge outputs. Specifically, the activated fine-grained edges should be consistent with the generic edges. 2). Such an alignment approach should take into account not only the pixel-wise features (intensity, etc.) but the spatial distance between the generic edges and fine-grained edges as well.





Hence, such a problem can be segregated into two distinct components, specifically the aggregation of fine-grained edge outputs and the alignment of fine-grained and generic edges. Concerning the former, the maximum operation suffices, as our objective is that the pixel characterized by a high response value in the generic edge output attains the highest possible response value, irrespective of the specific fine-grained edge output that causes its activation. The responsibility for determining which specific fine-grained edge output should induce this high response value lies with the supervised loss. Regarding the latter, our observation suggests that the inverse transformation network \cite{inverse-form-loss} satisfies the aforementioned requirements. Specifically, when trained on natural images with homography transformations, the inverse transformation network exhibits a robust capacity for spatial distance measurement.



As shown in \cref{fig:EA2Loss}, we first aggregate the four fine-grained edge outputs into one by pixel-wise maximum operation, obtaining $\mathcal{G}_{\rm rind}$. Moreover, an inverse transformation network $\Phi$ is used to predict the geometric alignment parameter between the generic edge output $\mathcal{E}^e$ and $\mathcal{G}_{\rm rind}$. If there is a perfect match between $\mathcal{E}^e$ and $\mathcal{G}_{\rm  rind}$, $\Phi$ should estimate an identity matrix. Thus, by estimating the distance between the output of $\Phi$ and the identity matrix and minimizing this distance, both the fine-grained and generic edges are able to acquire an explicit constraint. The formulation is expressed as follows:

\begin{equation}
\begin{aligned}
\mathcal{L}_{\rm alignment} &= || \hat\theta - I||_F ,\\
\hat\theta &= \Phi(\mathcal{E}^e,\mathcal{G}_{\rm rind}),\\
\mathcal{G}_{\rm rind} &= {\rm Maximum}(\mathcal{E}^r, \mathcal{E}^i, \mathcal{E}^n, \mathcal{E}^d).\\
\label{eq:EA2-loss}
\end{aligned}
\end{equation}
Here, $I$ is an identity matrix, and $||\cdot||$ indicates the Frobenius norm.

\subsection{Training}
\textbf{Loss function}:
We use the attention loss function presented in \cite{rindnet} to supervise the training of our fine-grained edge and generic edge, which can be formulated as:

\begin{equation}
\begin{aligned}
\mathcal{L}(\mathcal{Y},\mathcal{E})_{\rm erind} &= \sum_{k \in\{e,r,i,n,d\}} \lambda_k l(\mathcal{Y}^k,\mathcal{E}^k),\\
l(\mathcal{Y}^k,\mathcal{E}^k) &= -\sum_{i,j} (\mathcal{Y}_{i,j} \alpha\beta_k^{(1-\mathcal{E}_{i,j})^{\gamma_k}}\log(\mathcal{E}_{i,j})\\ &+  (1-\mathcal{Y}_{i,j})(1-\alpha)\beta_k^{\mathcal{E}_{i,j}^{ \gamma_k}}\log(1-\mathcal{E}_{i,j})).
\end{aligned}
\label{attention_loss_2}
\end{equation}

Here, $\mathcal{E}$ is the final prediction and $\mathcal{Y}$ is the corresponding ground truth label. $\lambda_k$ is a hyperparameter for task $k$. $\alpha = \frac{|\mathcal{Y}_-|}{|\mathcal{Y}|}$ and $1-\alpha = \frac{|\mathcal{Y}_+|}{|\mathcal{Y}|}$, where $\mathcal{Y}_-$ and $\mathcal{Y}_+$ indicate non-edge and edge ground truth labels, respectively. Furthermore, $\gamma_k$ and $\beta_k$ represent the hyperparameters associated with task $k$, while $\lambda_k$ serves as the balancing weight for this task. Thus, the total loss can be expressed as:
\begin{equation}
\mathcal{L}_{\rm total} = \mathcal{L}_{\rm erind}+\lambda_{a} \mathcal{L}_{\rm alignment},
\label{total_loss}
\end{equation}
where $\lambda_{a}$ is the balancing weight for alignment loss.


\section{Experiments}

\begin{table*}[th]
  \centering
  \caption{Quantitative comparison for REs, IEs, NEs, DEs, and Average on BSDS-RIND \cite{rindnet}. The values highlighted in red denote the performance margin in comparison to the second best method.}
  \resizebox{0.9\textwidth}{!}{
    \begin{tabular}{c|ccc|ccc|ccc|ccc|ccc}
    \toprule
    \multirow{2}{*}{Method} & \multicolumn{3}{c|}{Reflectance}& \multicolumn{3}{c|}{Illumination} & \multicolumn{3}{c|}{Normal}& \multicolumn{3}{c|}{Depth} & \multicolumn{3}{c}{Average} \\
          & ODS   & OIS   & AP    & ODS   & OIS   & AP    & ODS   & OIS   & AP    & ODS   & OIS   & AP    & ODS   & OIS   & AP \\
    \midrule
    HED\cite{hed}   & 0.412 & 0.466 & 0.343 & 0.256 & 0.290 & 0.167 & 0.457 & 0.505 & 0.395 & 0.644 & 0.679 & 0.667 & 0.442 & 0.485 & 0.393 \\
    RCF\cite{RCF}   & 0.429 & 0.448 & 0.351 & 0.257 & 0.283 & 0.173 & 0.444 & 0.503 & 0.362 & 0.648 & 0.679 & 0.659 & 0.445 & 0.478 & 0.386 \\
    DFF \cite{DFF}  & 0.447 & 0.495 & 0.324 & 0.290 & 0.337 & 0.151 & 0.479 & 0.512 & 0.352 & 0.674 & 0.699 & 0.626 & 0.473 & 0.511 & 0.363 \\
    RINDNet \cite{rindnet} & 0.478 & 0.521 & 0.414 & 0.280 & 0.337 & 0.168 & 0.489 & 0.522 & 0.440 & 0.697 & 0.724 & 0.705 & 0.486 & 0.526 & 0.432 \\
    EDTER \cite{transformer-based-edge-detector} & 0.496	& 0.552	& 0.440 &  0.341	& 0.363	& 0.222 & 0.513	& 0.557	& 0.459 & \textbf{0.703} & 0.733	& 0.695 & 0.513 & 0.551 & 0.454 \\
    \midrule
     ECT (Ours) & \textbf{0.520} & \textbf{0.567} & \textbf{0.470} & \textbf{0.371} & \textbf{0.399} & \textbf{0.318 (\color{red}{$\uparrow$} 9.6 \% }) & \textbf{0.516} & \textbf{0.558} & \textbf{0.473} & 0.699 & \textbf{0.734} & \textbf{0.722} & \textbf{0.526} & \textbf{0.564} & \textbf{0.496} \\
    \bottomrule
    \end{tabular}%
    }
  \label{tbl:comparison_to_sota}
\end{table*}

\begin{table*}[th]
  \centering
  \caption{Ablation study to verify the effectiveness of stage-wise adaptive learned cause tokens and EA2 Loss on BSDS-RIND \cite{rindnet}.}
  \resizebox{1\textwidth}{!}{
    \begin{tabular}{cc|ccc|ccc|ccc|ccc|ccc}
    \toprule
    \multicolumn{2}{c|}{Method}    & \multicolumn{3}{c|}{Reflectance} & \multicolumn{3}{c|}{Illumination} & \multicolumn{3}{c|}{Normal} & \multicolumn{3}{c|}{Depth} & \multicolumn{3}{c}{Average} \\
     \makecell[c]{Stage-wise Adaptive \\ Learned Cause Tokens} & \makecell[c]{EA2 Loss}& ODS   & OIS   & AP    & ODS   & OIS   & AP    & ODS   & OIS   & AP    & ODS   & OIS   & AP    & ODS   & OIS   & AP \\
    \midrule
  \XSolidBrush & \XSolidBrush & 0.508 & 0.532 & 0.452 & 0.355 & 0.384 & 0.250 & 0.503 & 0.546 & 0.440 & \textbf{0.704} & \textbf{0.736} & 0.692 & 0.518 & 0.549 & 0.459 \\
        \XSolidBrush    & \checkmark     & 0.511 & 0.540 & 0.448 & 0.367 & 0.386 & 0.292 & 0.503 & 0.546 & 0.443 & 0.699 & 0.733 & 0.706 & 0.520 & 0.551 & 0.472 \\
        \checkmark    &   \XSolidBrush    & 0.519 & 0.562 & 0.466 & 0.368 & 0.398 & 0.305 & 0.511 & 0.552 & 0.462 & \textbf{0.704} & \textbf{0.736} & 0.719 & 0.525 & 0.562 & 0.488 \\
          \checkmark    & \checkmark     & \textbf{0.520} & \textbf{0.567} & \textbf{0.470} & \textbf{0.371} & \textbf{0.399} & \textbf{0.318} & \textbf{0.516} & \textbf{0.558} & \textbf{0.473} & {0.699} & {0.734} & \textbf{0.722} & \textbf{0.526} & \textbf{0.564} & \textbf{0.496} \\
        \bottomrule
    \end{tabular}
    }
  \label{tab:ablation_table}
\end{table*}

\begin{table}[th]
    \centering
    \caption{Backbone and network parameters.}
    \resizebox{\linewidth}{!}{
    \begin{tabular}{c|c|c}
    \toprule
    Method & Backbone & Parameters \\
    \midrule
    HED \cite{hed} & VGG-16 & 14 M (14,720,620) \\
    RCF \cite{RCF} & VGG & 14 M (14,804,129) \\
    DFF \cite{DFF} & ResNet-50 & 25 M (25,669,517) \\
    RINDNet \cite{rindnet} & ResNet-50 & 59 M (59,388,357) \\
    EDTER \cite{transformer-based-edge-detector} & ViT-Large + ViT-Base & 468 M (468,780,434)\\
    ECT (Ours) & ViT-Hybrid & 145 M (145,283,659) \\
    \bottomrule
    \end{tabular}
    \label{tab:main-parameter-number-comparison}
    }
\end{table}


\begin{figure}[th]
\centerline{\includegraphics[width=1\columnwidth]{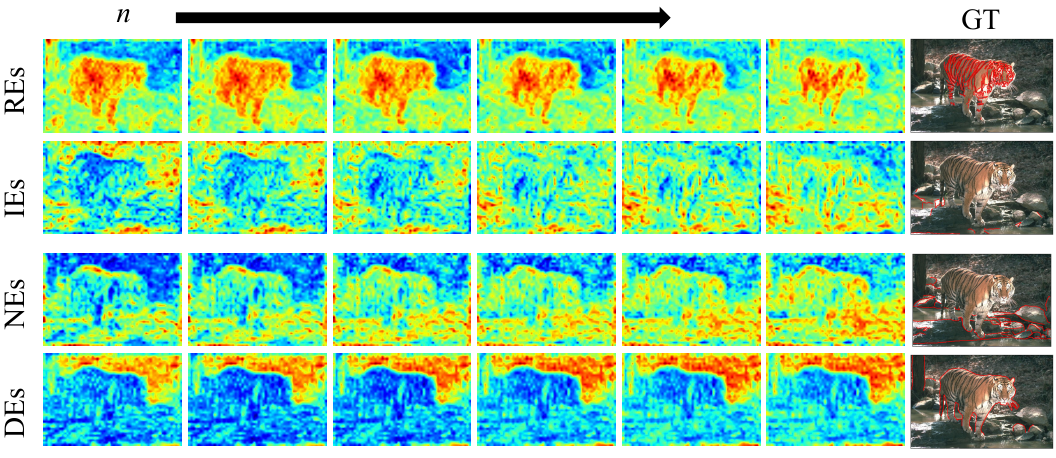}}
\caption{Visualization of attention maps generated by the cause-aware decoder. The decoder stage progresses from left to right, while the attention maps for different edge types, namely REs, IEs, NEs, and DEs, are displayed from top to bottom}
\label{fig:cause-aware decoder attention}
\vspace{-0.5em}
\end{figure}


\subsection{Datasets and Evaluation Metrics}

\textbf{Dataset:} 
\label{section-part: bsds-dataset definition}
BSDS-RIND \cite{rindnet} is a re-labeled dataset for fine-grained edge detection based on BSDS \cite{metrics}, consisting of 500 annotated images with 2,595,344 edge pixels including 663,039 REs, \textbf{210,711 IEs}, 685,424 NEs and 1,036,170 DEs. We follow \cite{rindnet}'s data augmentation process, train and test splits, and evaluation procedure for a fair comparison.

\textbf{Evaluation metrics:} Consistent with \cite{rindnet}, we utilized three standard evaluation metrics, namely the F1 score under a fixed contour threshold (ODS), the F1 score under the per-image best threshold (OIS), and the average precision (AP). 
\label{subset:bsds-rind matrics}

\subsection{Comparisons with State-of-the-art Methods}

We conducted a comprehensive comparison of our method with state-of-the-art edge detectors, namely RINDNet \cite{rindnet}, HED \cite{hed}, RCF \cite{RCF}, DFF \cite{DFF}. These results are obtained from \cite{rindnet}. In addition, to demonstrate the effectiveness of our method, we reproduce another transformer-based model, EDTER \cite{transformer-based-edge-detector}. To enable EDTER to handle multiple edge types simultaneously, we modify its output from $\mathcal{E} \in \{0, 1\}^{W \times H}$ to $\mathcal{E} \in \{0, 1\}^{4 \times W \times H}$ following RINDNet \cite{rindnet}.

To ensure fairness in the experiments, we collected information on the network parameter numbers of different methods, as shown in \cref{tab:main-parameter-number-comparison}.  

\begin{figure*}[th]
\centerline{\includegraphics[width=1\textwidth]{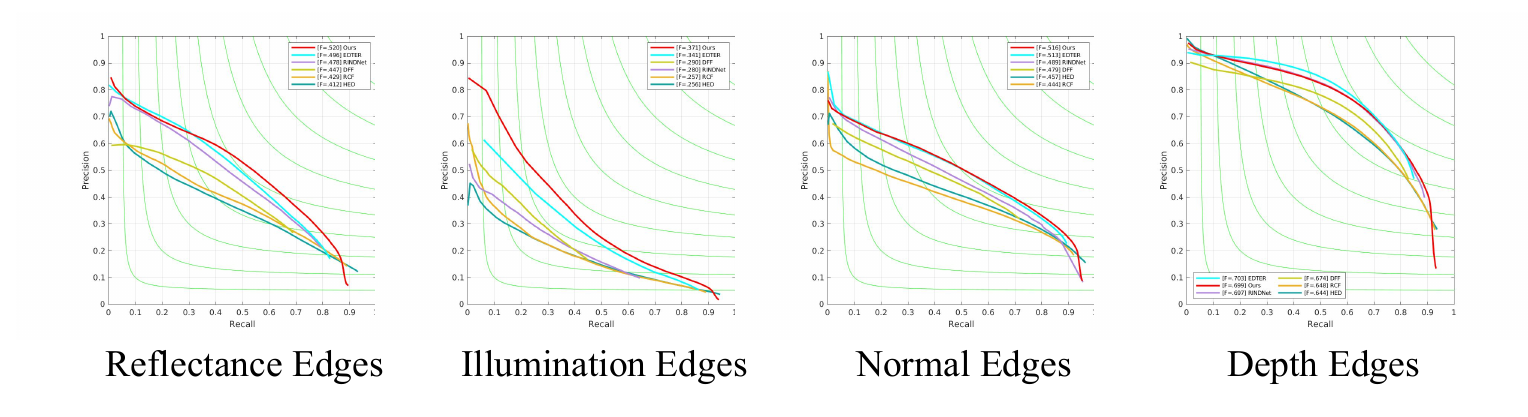}}
\caption{Precision-recall curves for four fine-grained edge types. A better method would appear on the right-top part.}
\label{fig:precision and recall curve}
\end{figure*}

\textbf{Quantitative comparison:} As demonstrated in \cref{tbl:comparison_to_sota} and \cref{fig:precision and recall curve}, our method exhibits significant improvements over the previous state-of-the-art (SOTA) approaches. Remarkably, our method achieves an average precision score of 0.318 in the IEs category, which surpasses the second- and third-best performers, transformer-based detector EDTER (0.222) and RCF (0.173), by 9.6\% and 14.5\%, respectively. These results further demonstrate that our superiority is not solely attributable to the transformer-based architecture, but also to the two-stage design and learned cause token.

\textbf{What are the reasons behind the significant improvement observed in IE:} Sample number of IE is significantly less than that of all other edge types \cite{rindnet}. We believe that this imbalance in sample numbers presents a unique challenge for IE, as evidenced by the small gain observed from using a larger network in the comparison between HED and RINDNet (see \cref{tbl:comparison_to_sota} and \cref{tab:main-parameter-number-comparison}). We attribute this phenomenon to the data-hungry nature of previous methods, which implies that the gain would increase with more IE samples.

In contrast, ECT explicitly models the IE cause as learned cause tokens, which we argue is less data-hungry. In other words, even with limited samples, ECT can still learn a good prior of edge causes, leading to better performance,
which can also be proven by the \cref{fig:cause-aware decoder attention} where attention maps for each cause token are shown to focus on different pixels based on the corresponding edge cause.

\begin{figure*}[th]
\centerline{\includegraphics[width=1\textwidth]{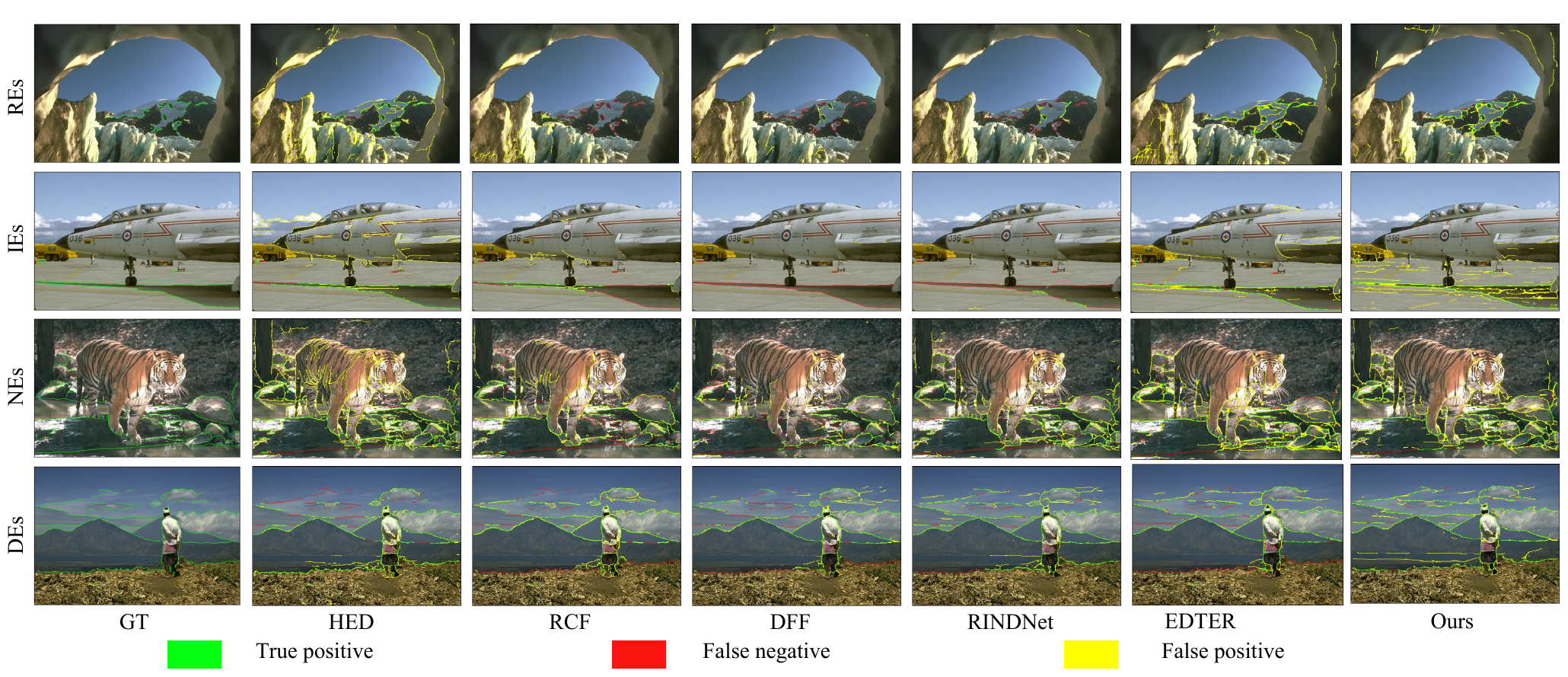}}
\caption{Qualitative comparison for REs, IEs, NEs and DEs on BSDS-RIND \cite{rindnet}.}
\label{fig:qualitative}
\end{figure*}

\textbf{Qualitative comparison:} As depicted in \cref{fig:qualitative}, ECT exhibits a reduced number of false negatives in comparison to prior SOTA methods, thus providing further evidence of its superiority.



\begin{figure}[th]
\centerline{\includegraphics[width=1\columnwidth]{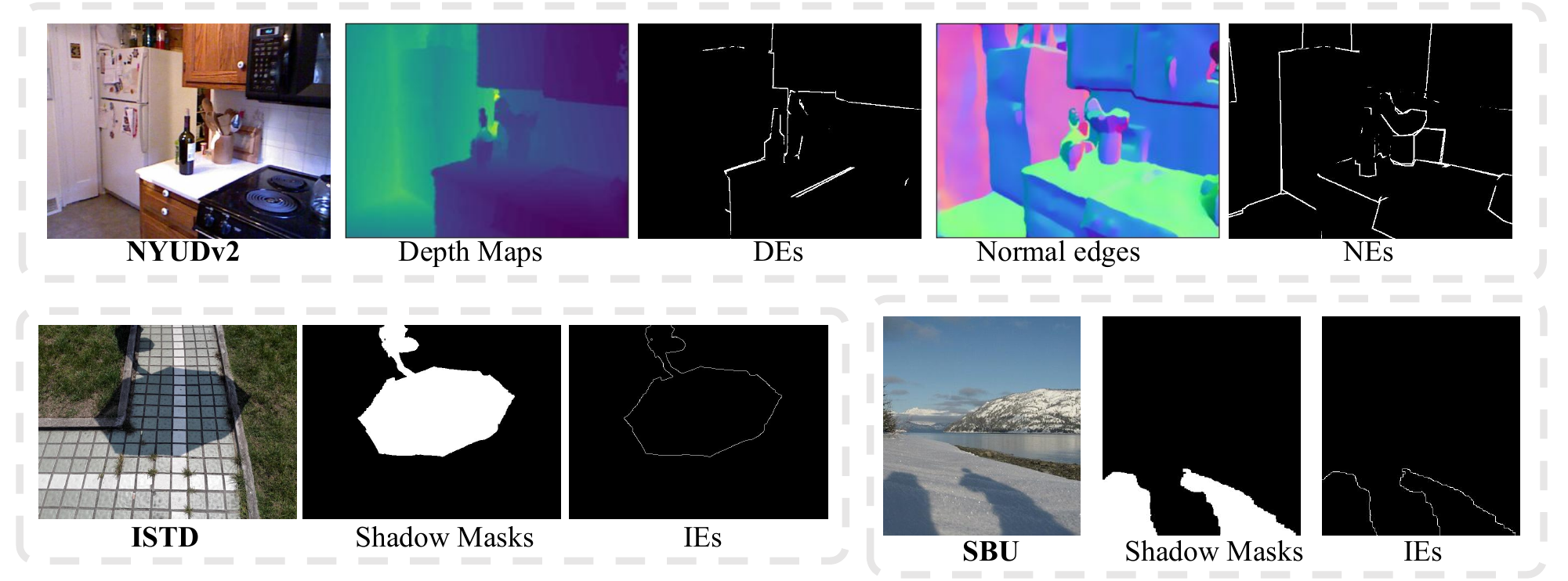}}
\caption{Qualitative results for newly generated datasets.}
\label{fig:new-dataset-qualitative-resutls}
\end{figure}

\begin{figure*}[th]
\centerline{\includegraphics[width=\textwidth]{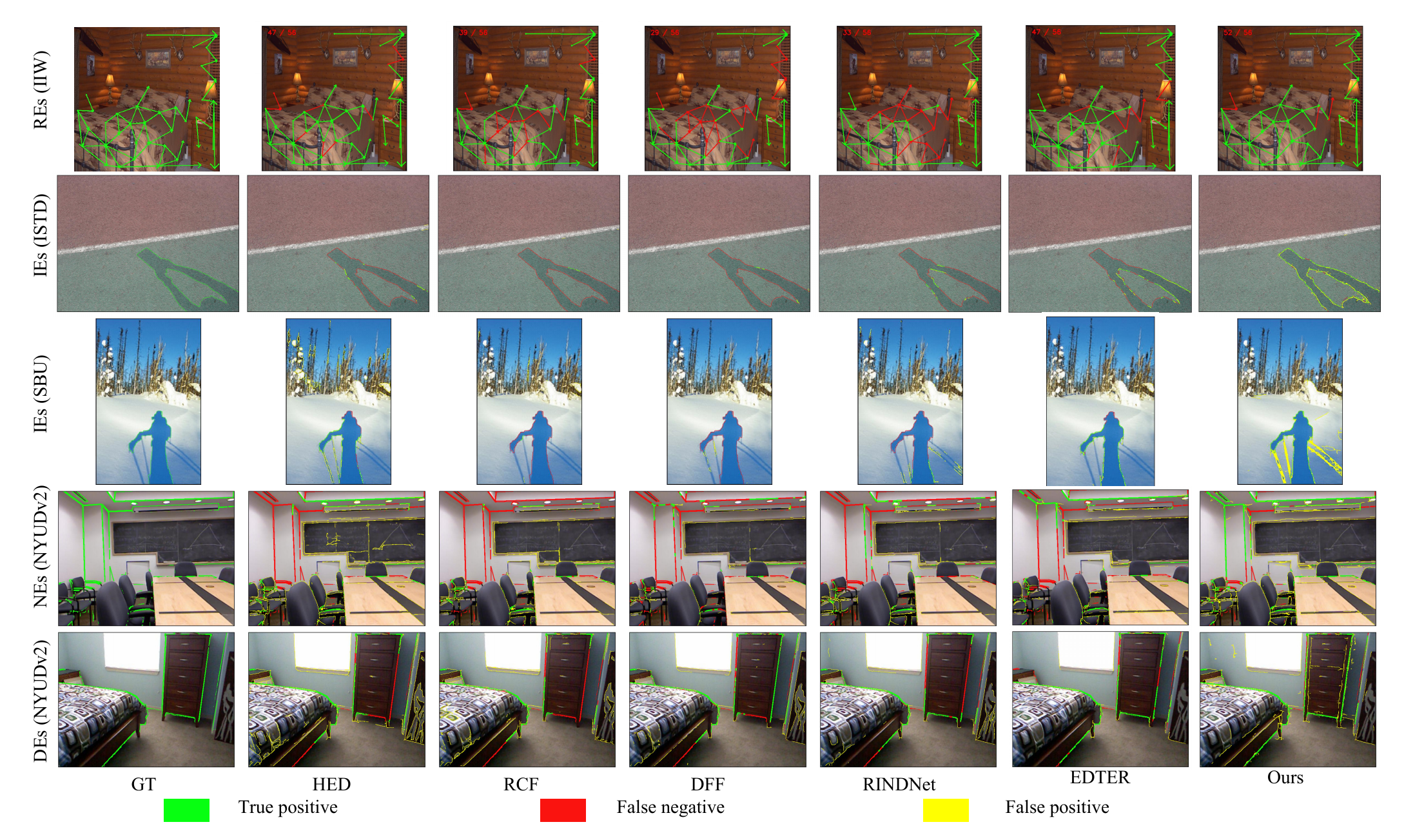}}
\caption{Qualitative comparison for REs (IIW \cite{bell2014intrinsic}), IEs (ISTD\cite{wang2018stacked}), IEs (SBU\cite{vicente2016large}), NEs (NYUDv2\cite{silberman2012indoor}), DEs (NYUDv2 \cite{silberman2012indoor})}
\label{fig:qualitative_comparison_for_new_dataset}
\end{figure*}

\subsection{Ablation Study}


In this section, we investigate the impact of the stage-wise adaptive learned cause tokens and EA2 loss. The first row of \cref{tab:ablation_table} indicates the model that incorporates the inadaptive learned cause token but does not employ the EA2 Loss.

\textbf{Effectiveness of EA2 loss: } As shown in the third row of \cref{tab:ablation_table}, excluding the EA2 loss lead to a direct performance drop of 0.8\% in the average AP. This finding is consistent with the qualitative results presented in \cref{fig:loss_qualitative_results}. Without the EA2 constraint loss, both the depth and normal edges adjacent to the penguin's feet (see \cref{fig:loss_qualitative_results} left) and the reflectance edge on the road (see \cref{fig:loss_qualitative_results} right) are lost, despite being included in the outputs of the first stage. This suggests that the EA2 loss plays a crucial role in preserving the fine-grained edges in the second stage. In contrast, the method utilizing the EA2 loss explicitly accounts for these missing edges, further demonstrating the effectiveness of the EA2 loss in achieving superior edge detection performance.

\textbf{Effectiveness of stage-wise adaptive learned cause tokens: } Our model's performance is impacted by the exclusion of stage-wise adaptive learned cause tokens, as evidenced by the 2.4\% decrease in average AP. To further demonstrate the effectiveness of these tokens, we analyze the attention maps shown in \cref{fig:cause-aware decoder attention}. These maps are task-specific and highlight distinct regions of the image to justify fine-grained edges. For example, in the case of depth edges caused by depth discontinuity, accurate prediction requires significant attention to the background. As shown in \cref{fig:cause-aware decoder attention}, the attention map for depth causes primarily focuses on the background, and the response value increases as the decoder stages progress. The other attention maps share the same responsibility of attending to the regions of the image that are relevant to their respective causes.

Furthermore, when we removed both the EA2 loss and Stage-wise Adaptive Learned Cause Tokens, the performance further dropped by 3.7\% in average AP. This result highlights the crucial role of these two components in our model's edge detection performance.



\subsection{Transferability Experiments}

In order to showcase the transferability of our method, we conducted additional comparison experiments on multiple datasets including \textbf{IIW \cite{bell2014intrinsic} (REs), SBU \cite{vicente2016large}, ISTD \cite{wang2018stacked} (IEs) and NYUDv2 \cite{silberman2012indoor} (NEs and DEs)}. Importantly, no additional training was performed on these datasets; instead, all comparison models were trained exclusively on BSDS-RIND and subsequently validated on the test set of the aforementioned datasets. Through this rigorous evaluation, we are able to demonstrate the robustness and generalization ability of our proposed approach across a range of diverse scenarios and out-of-domain datasets.

\subsubsection{Introduce to New Dataset and Benchmark}

The \textbf{IIW} test set comprises 1046 images that have been annotated in the form of pairwise reflectance comparisons between two distinct points, with a specific focus on decomposing the intrinsic component of the images. We assume that an RE exists if the reflectance value of the point pair is not equal.

The \textbf{SBU} and \textbf{ISTD} datasets are two commonly used datasets for shadow detection, both of which provide shadow mask maps and contain 638 and 540 images in their respective test sets. We generate the IEs through local variation intensity from the shadow mask maps. The detailed process for generating the IEs can be found in the supplementary materials. The qualitative results of the newly generated dataset are shown in \cref{fig:new-dataset-qualitative-resutls}. 

The \textbf{NYUDv2} contains 654 RGB-D data of 464 different indoor scenes with detailed annotations, including depth maps, surface normal maps, semantic segmentation labels, and object instance labels. We generated the GEs from the instance labels following \cite{gupta2013perceptual}. Furthermore, the DEs and NEs are generated from GEs by evaluating the local variation intensity in the depth and normal maps. The detailed process for generating the NEs and DEs can be found in the supplementary materials. Moreover, as depicted in \cref{fig:new-dataset-qualitative-resutls}, the generated normal and depth edges are not perfect as the normal and depth maps contain imperfections, particularly in the regions adjacent to the edges in the depth and normal maps, which are mainly caused by limitations of the data collection device. However, it's trivial since the fairness of the comparison is ensured.

We adopted the BSDS-RIND metric \cite{rindnet} for evaluation of the SBU, ISTD, and NYUDv2 datasets. For the IIW dataset, we measure the mean recall under the threshold range of 0.01 to 0.99, with a step of 0.01.

\subsubsection{Transferability Comparisons}
As demonstrated in \cref{Tab:quantitative results under-IIW-ISTD-SBU-NYUDv2}, our proposed method achieves promising results across different datasets. Particularly noteworthy is the reduction in the performance gap in the estimation of IEs, where the margin from the second best method in average precision (AP) drops from 9.6\% to 3.4\% in SBU. We attribute this reduction to the noise present in the SBU dataset. As an example, \cref{fig:qualitative_comparison_for_new_dataset} shows that some illumination edges of ski poles are missing in the ground truth annotations of the SBU dataset.

\begin{table*}[t]
  \centering
  \caption{ Quantitative comparison for REs (IIW \cite{bell2014intrinsic}), IEs (ISTD\cite{wang2018stacked}), IEs (SBU\cite{vicente2016large}), NEs (NYUDv2\cite{silberman2012indoor}), DEs (NYUDv2 \cite{silberman2012indoor}). The values highlighted in red denote the performance margin in comparison to the second best method.}
  \resizebox{0.9\textwidth}{!}{
    \begin{tabular}{c|c|ccc|ccc|ccc|ccc}
    \toprule
    \multirow{2}{*}{Method} & \multicolumn{1}{c|}{Reflectance (IIW \cite{bell2014intrinsic}) }& \multicolumn{3}{c|}{Illumination (ISTD \cite{wang2018stacked})}& \multicolumn{3}{c|}{Illumination (SBU \cite{vicente2016large})} & \multicolumn{3}{c|}{Normal (NYUDv2 \cite{silberman2012indoor})}& \multicolumn{3}{c}{Depth (NYUDv2 \cite{silberman2012indoor}) }\\
          & Mean Recall   & ODS   & OIS   & AP    & ODS   & OIS   & AP    & ODS   & OIS   & AP    & ODS   & OIS   & AP \\
    \midrule
    HED\cite{hed}   & 0.638 & 0.508 & 0.515 & 0.499  & 0.566 & 0.618 & 0.565 & 0.332 & 0.342 & 0.149 & 0.360 & 0.376 & 0.185 \\
    RCF\cite{RCF}   &  0.594 & 0.492 & 0.510 & 0.463 & 0.535 & 0.586 & 0.510 & 0.320 & 0.325 & 0.120 & 0.347 & 0.364 & 0.172 \\
    DFF \cite{DFF}  & 0.481 & 0.478 & 0.495 & 0.299 & 0.475 & 0.483 & 0.297 & 0.271 & 0.272 & 0.081 & 0.340 & 0.348 & 0.142 \\
    RINDNet \cite{rindnet} & 0.519 & 0.547 & 0.584 & 0.465 & 0.557 & 0.595 & 0.471 & 0.333 & 0.337 & \textbf{0.156} & 0.357 & 0.369 & 0.175 \\
    EDTER \cite{transformer-based-edge-detector} & 0.458 & 0.552 & 0.631 & 0.511 & 0.599 & 0.651 & 0.534 & 0.333 & 0.340 & 0.131 & 0.349 & 0.360 & 0.170 \\
    \midrule
     Ours & \textbf{0.641} & \textbf{0.642} & \textbf{0.689} & \textbf{0.664} & \textbf{0.591} & \textbf{0.656} & \textbf{0.599 (\color{red}$\uparrow$ 3.4\%}) & \textbf{0.343} & \textbf{0.352} & 0.146 & \textbf{0.369} & \textbf{0.383} & \textbf{0.197} \\
    \bottomrule
    \end{tabular}
    }
    \label{Tab:quantitative results under-IIW-ISTD-SBU-NYUDv2}
\end{table*}

\section{Conclusion}

In this paper, we design a two-stage transformer-based network to bridge the relationship between the fine-grained and generic edge detection tasks. The edge cause is modeled as four learnable tokens in a cause-aware decoder design. Moreover, an EA2 loss is proposed to make fine-grained and generic edge output more consistent. In addition to the BSDS-RIND benchmark, we conduct extensive experiments on several newly collected datasets. The experimental results show that our method achieves state-of-the-art performance on all the benchmark datasets.









\bibliographystyle{unsrt}
\bibliography{egbib}



\end{document}